\pdfoutput=1

\documentclass[10pt,twocolumn,letterpaper]{article}

\usepackage[pagenumbers]{iccv} 

\definecolor{iccvblue}{rgb}{0.21,0.49,0.74}
\usepackage[pagebackref,breaklinks,colorlinks,allcolors=iccvblue]{hyperref}

\usepackage{multirow}
\usepackage{booktabs}
\usepackage{caption}
\usepackage{tabularx}
\usepackage{graphicx}
\usepackage{array}
\usepackage{colortbl}
\usepackage{algorithm}  
\usepackage{algorithmic}
\usepackage{setspace}
\usepackage{marvosym}
\usepackage[final]{changes} 
\let\oldreplace\replaced
\renewcommand{\replaced}[2]{\oldreplace{#1}{\hspace{0.5em}#2}}

\setdeletedmarkup{\textcolor{red!30}{\sout{#1}}}
\setaddedmarkup{\textcolor{red}{#1}}

\def\paperID{xxxx} 
\def\confName{ICCV}
\def\confYear{2025}

\title{MIDAS: Modeling  Ground-Truth Distributions with  Dark \\ Knowledge for Domain Generalized Stereo Matching}

\author{Peng Xu \quad Zhiyu Xiang\thanks{Corresponding author.} \quad Jingyun Fu \quad Tianyu Pu \quad Hanzhi Zhong \quad Eryun Liu\\
Zhejiang University, China\\
\url{https://github.com/xxxupeng/MIDAS}
}

\begin{document}

\maketitle
\begin{abstract}
\replaced{Despite the significant advances in domain generalized stereo matching, existing methods still exhibit domain-specific preferences when transferring from synthetic to real domains, hindering their practical applications in complex and diverse scenarios. The probability distributions predicted by the stereo network naturally encode rich similarity and uncertainty information. Inspired by this observation, we propose to extract these two types of dark knowledge from the pre-trained network to model intuitive multi-modal ground-truth distributions for both edge and non-edge regions. To mitigate the inherent domain preferences of a single network, we adopt network ensemble and further distinguish between objective and biased knowledge in the Laplace parameter space. Finally, the objective knowledge and the original disparity labels are jointly modeled as a mixture of Laplacians to provide fine-grained supervision for the stereo network training. Extensive experiments demonstrate that: (1) Our method is generic and effectively improves the generalization of existing networks. (2) PCWNet with our method achieves the state-of-the-art generalization performance on both KITTI 2015 and 2012 datasets. (3) Our method outperforms existing methods in comprehensive ranking across four popular real-world datasets.}{
Existing stereo networks exhibit the specific cross-domain preferences in generalizing from synthetic to real domains, hindering their practical applications in complex and diverse scenes. In this paper, we deal with this problem by leveraging the dark knowledge within the ensemble models to generate more reasonable and insightful supervision signals for stereo matching. The multi-modal distributions output by pre-trained stereo networks intrinsically capture rich information, such as matching similarity and uncertainty. To aggregate the objective knowledge and filter out the biased knowledge within the rich information, we propose to project the modes of the output distributions into the Laplace parameter space, and then cluster and fuse these modes. Finally, we model the ground-truth distribution as a mixture of Laplacians for training stereo matching networks. Extensive experiments on different stereo architectures demonstrate the generality and effectiveness of our method. Our method outperforms existing methods in generalization performance on the KITTI 2015 and KITTI 2012 datasets. Meanwhile, we ranks $1^{st}$ in the comprehensive comparison on four popular datasets.
}
\end{abstract}    
\section{Introduction}
\label{sec:intro}
Stereo matching \replaced{plays a crucial role in numerous}{is widely used in many} vision-based applications, \replaced{including}{such as} automatic driving, augmented reality, and robotics. \replaced{For}{In} these safety-critical systems, \added{ensuring} the reliability of stereo networks \replaced{in}{in dealing with} complex \replaced{real-world scenarios}{and diverse scenes} \replaced{is of paramount importance}{cannot be overstated}. \replaced{Different from monocular depth estimation~\cite{depthanything}, collecting massive amounts of labeled or even unlabeled real data to train a stereo foundation model is extremely challenging due to the hardware limitations.}{An ideal solution is to collect a large amount of real data covering a variety of scenes for training, such as Depth Anything~\cite{depthanything}. However, the real stereo data is not as easy to obtain as monocular data due to the hardware limitations.} Existing works~\cite{DSMNet,ITSA,GraftNet,hvt, adl} attempt to directly train \added{well-generalizing stereo} networks in the synthetic domain~\cite{SceneFlow} \deleted{and generalize to the real domain,} \replaced{and}{which} have achieved \replaced{considerable progress}{remarkable results}.

\begin{figure}[t]
    \centering
    \includegraphics[width=1.0\linewidth]{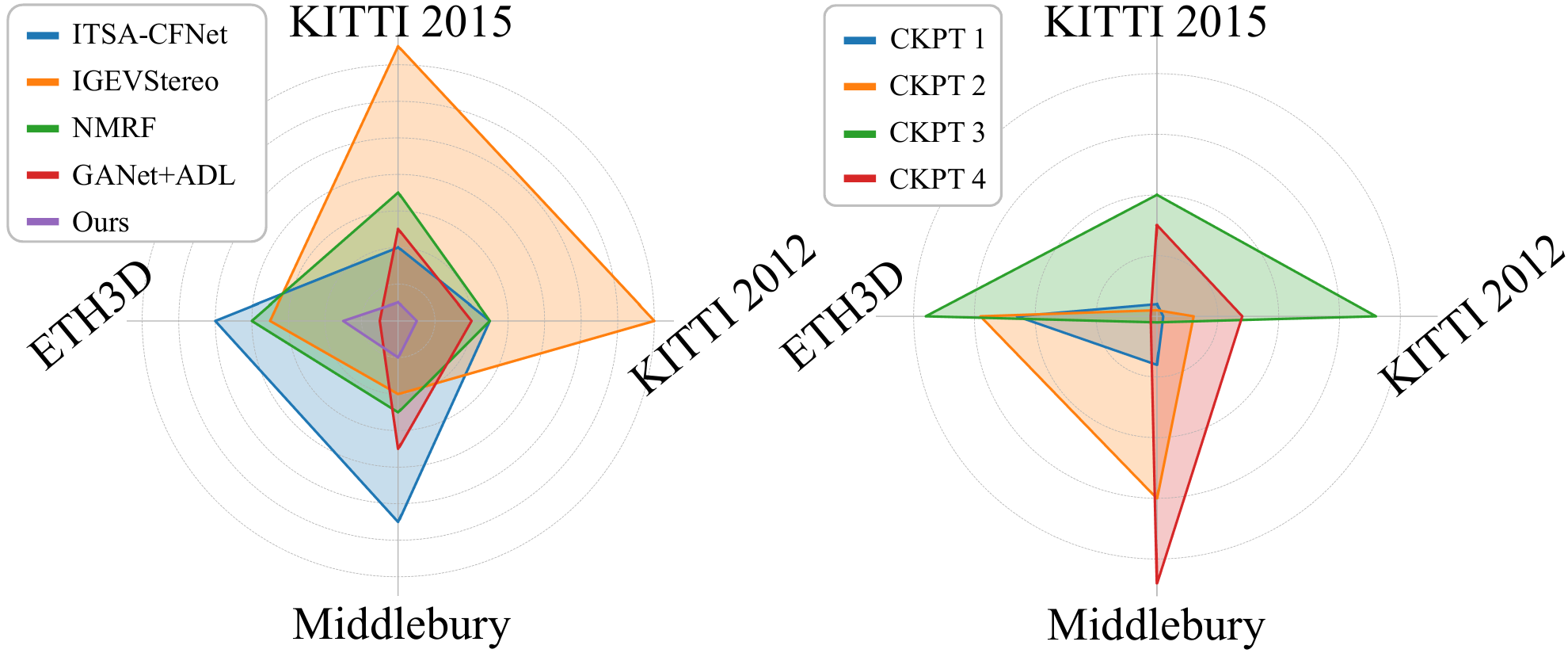}
    \caption{\textbf{Cross-domain preferences} of different network architectures (left) and different checkpoints of the same network~\cite{adl} (right). All the methods are trained on synthetic  dataset~\cite{SceneFlow} and evaluated on \added{four} real-world datasets~\cite{KITTI2012,KITTI2015,Middlebury,ETH3D}. The closer to the center, the better the performance.}
    \label{radar}
\end{figure}

Nevertheless, these \replaced{methods}{well-generalized networks} still exhibit undesired cross-domain preferences. As shown in~\cref{radar} (left), ITSA-CFNet~\cite{ITSA} performs well on \deleted{the} KITTI 2015~\cite{KITTI2015} and 2012~\cite{KITTI2012}\deleted{,} but poorly on \deleted{the} Middlebury~\cite{Middlebury} and ETH3D~\cite{ETH3D}\replaced{, while IGEVStereo~\cite{igev} is just the opposite.}{ IGEVStereo~\cite{igev} shows the opposite trend.} \deleted{In addition,} Rao~\etal\cite{maskedstereo} point out that existing works \replaced{typically report the optimal result among all checkpoints for each dataset separately. This conflicts with the practical demands, where a single network with fixed weights must maintain robust performance across diverse scenarios.}{prefer to report the best results on each dataset, which are usually from the different checkpoints. In practical applications, the deployed deep network can only load one fixed checkpoint to handle diverse scenes.} 
\replaced{Motivated by these limitations, our work aims to mitigate}{Given this, we seek to alleviate} the domain preference\added{s} and improve the \replaced{comprehensive}{cross-domain} generalization \replaced{of}{with} a single checkpoint\deleted{ model}.

\begin{figure*}
    \centering
    \includegraphics[width=0.98\linewidth]{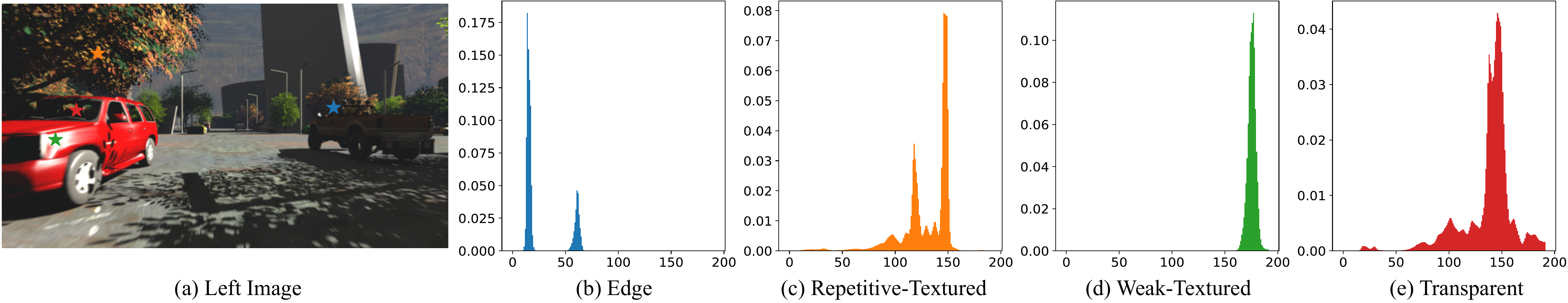}
    \vspace{-2pt}
    \caption{\textbf{Visualization of the probability distributions} output by PSMNet~\cite{PSMNet}, \deleted{which is} trained with \deleted{the} uni-modal Laplacian cross-entropy loss~\cite{PDSNet}. \replaced{Four representative pixels (marked with colored stars in the left image) illustrate different distribution patterns.}{Four sampling pixels marked with stars in the left image are taken as examples.} \deleted{It can be found that} \replaced{T}{t}he stereo network \replaced{naturally produces}{tends to output} multi-modal distributions in edge \textcolor[HTML]{1F77B4}{$\bigstar$}, repetitive-textured \textcolor[HTML]{ff7f0e}{$\bigstar$}, and transparent \textcolor[HTML]{d62728}{$\bigstar$} regions, \replaced{while generating}{and output} wider \replaced{uni-modal distribution}{mode} in weak-textured \textcolor[HTML]{2ca02c}{$\bigstar$} region\deleted{s}.}
    \label{distribution visualization}
    \vspace{-8pt}
\end{figure*}


\replaced{Guiding stereo networks to learn fine-grained disparity distributions has been proven effective in enhancing cross-domain generalization.}{We achieve this goal by providing the more intuitive and insightful ground-truth distributions for stereo network training.} Currently, the \added{stereo} ground-truth \deleted{disparity distribution in stereo matching} is usually modeled as the uni-modal Laplacian or Gaussian distribution\added{, centered around the disparity label}~\cite{PDSNet,SMNet,AcfNet}. \replaced{Xu~\etal~\cite{adl} model the multi-modal ground-truth distributions for edge regions but retain uni-modal modeling for non-edge regions, which limits their ability to address repetitive- or weak-textured regions. \deleted{Furthermore,~}\cref{radar} (right) demonstrates that their different checkpoints also exhibit varying domain preferences.}{Xu~\etal~\cite{adl} further extract edge information from the disparity maps to construct the multi-modal ground-truth, which greatly improves the generalization of the stereo networks.}


\deleted{In this paper, we aim to unleash the power of dark knowledge from pre-trained stereo networks in modeling stereo ground-truth distributions. }Our exploration stems from the observation that stereo matching networks naturally \deleted{prefer to} output multi-modal distributions \added{not only in edge region but also} in \added{non-edge} regions with repetitive textures\deleted{, edges,} or transparent surfaces, as shown in~\cref{distribution visualization}. \replaced{These}{The} multi-modal \replaced{distributions}{outputs} in stereo matching \replaced{are analogous to}{is similar to} the multi-hot \replaced{distributions}{outputs} in classification \added{task~\cite{hinton2015distilling}, with both encoding the similarity information}. \deleted{Guiding classification networks to learn the dark knowledge output from pre-trained networks has been shown effective in enhancing the generalization performance of the supervised learners~\cite{hinton2015distilling}. We believe that the multi-modal outputs of the stereo matching network implicitly contain rich dark knowledge, which can be extracted and used to supervise network training.} \replaced{However, stereo matching differs from classification in the distribution patterns. While classification networks produce responses characterized by Dirac delta functions at discrete class locations, stereo matching networks generate modes with finite width for each correspondence candidate. This width encodes valuable uncertainty information about the pixel-wise matching process.}{In addition, stereo matching networks differ from classification networks as their output distributions are multiple modes rather than multiple Dirac delta functions. The width of the mode can reflect the uncertainty of the match.} When matching becomes \replaced{highly uncertain}{more difficult}, stereo networks \deleted{can} output \replaced{the}{a} wider mode\added{s, as shown in~\cref{distribution visualization} (d)}.

\added{Building upon the insights into the properties of stereo distributions, we propose to extract dark knowledge \added{(similarity and uncertainty)} from the pre-trained stereo network to model informative ground-truth distributions for both edge and non-edge regions.}
\replaced{W}{Specifically, w}e \replaced{adopt}{refer to the} \replaced{network ensemble}{ensemble models}~\cite{dk_ensemble1,dk_ensemble2,dk_ensemble3}  to mitigate the \replaced{domain}{undesired} preferences inherent in \replaced{a single}{the} pre-trained network\deleted{s}. However, \replaced{na\"ively}{directly} superimposing the output\replaced{s from}{ distributions of the} the ensemble \replaced{fails to effectively fuse the uncertainty information and compromises the uni-modal nature}{ models may harm the unimodality} of each mode. To address \replaced{these problems}{this issue}, we \replaced{innovatively aggregate the}{propose to fuse} dark knowledge in the Laplace parameter space.  \replaced{Specifically}{First}, we \added{first} separate \added{individual} modes from the \added{output} multi-modal distributions and project them \replaced{as a series of points in}{into} the Laplace \added{parameter} space. Then, we cluster \replaced{these points}{the modes in the parameter space} to distinguish \replaced{the objective knowledge (effective clusters)}{the effective clusters (objective knowledge)} from \replaced{the biased knowledge (noise)}{the noise (biased knowledge)}. Finally, the \replaced{objective}{aggregated} knowledge, together with the disparity labels, \replaced{is utilized}{are used} to construct the multi-modal \added{ground-truth} distribution\replaced{s}{ of the ground-truth}.

We conduct extensive experiments to demonstrate the effectiveness of our method \added{on PSMNet~\cite{PSMNet}, GwcNet~\cite{GwcNet}, and PCWNet~\cite{PCWNet}}. \replaced{At}{As} the time of writing this paper, PCWNet with our method \replaced{achieves the state-of-the-art generalization performance among all published methods}{ranks $1^{st}$} on \added{both} \deleted{the} KITTI 2015~\cite{KITTI2015} and 2012~\cite{KITTI2012} datasets. \replaced{For a more intuitive comparison of comprehensive generalization performance, we compute the mean rank across four real-world datasets}{To intuitively compare the comprehensive generalization performance in diverse scenes, we further rank our method and the competitors based on the average of rankings on four popular real-world datasets}~\cite{KITTI2012,KITTI2015,Middlebury,ETH3D}. \replaced{PCWNet}{All three backbones~\cite{PSMNet,GwcNet,GANet}} with our method \added{also} outperform\added{s} \replaced{the existing}{previous} \deleted{state-of-the-art} methods.

Our contributions can be summarized as follows: 

\begin{itemize}
    \item To the best of our knowledge, we are the first to model ground-truth distributions with adaptive mode number and width for \added{all regions in} stereo matching\deleted{ using dark knowledge within ensemble models}. \added{By embedding similarity and uncertainty information into modeling, we provide intuitive and insightful supervision for network learning.}
    \item \replaced{We propose to aggregate objective knowledge in the Laplace parameter space, reducing the introduction of biased knowledge during the modeling process.}{We propose to aggregate the dark knowledge in the Laplace parameter space and effectively filter out the biased knowledge, which is harmful to network learning.}
    \item We achieve the state-of-the-art cross-domain generalization performance on KITTI 2015 and 2012 datasets, and rank $1^{st}$ in the comprehensive comparison \replaced{across}{of} four popular real-world datasets. 
    \item  Our method \added{achieves stable cross-domain generalization across multiple datasets with the fixed checkpoint and} demonstrates \replaced{superior}{excellent} robustness \replaced{when handling}{in dealing with} various challenging regions, \replaced{significantly enhancing the reliability of}{which can strongly guarantee the safety of} stereo vision systems in real-world applications.
\end{itemize}


\section{Related Work}

\textbf{Deep Stereo Matching.} \deleted{Stereo matching has been studied for several decades.} In recent years, deep learning-based \added{stereo matching} methods \added{have} become mainstream and achieve performance far exceeding traditional methods. DispNet~\cite{SceneFlow} is the first end-to-end stereo network. GCNet~\cite{GCNet} aggregates information through 3D convolution. PSMNet~\cite{PSMNet} and GwcNet~\cite{GwcNet} are the two most popular stereo \replaced{backbones}{baselines}~\cite{AcfNet,GraftNet,ITSA,ACVNet,FCNet,SMDNet,adl,chen2022normalized}, prompting the prosperity of the community. The former improves the extracted features through \deleted{the} spatial pyramid pooling~\cite{spp}\added{,} while the latter preserves \deleted{the} feature similarity information in the cost volume through the group-wise correlation. Besides the cost volume filtering-based methods mentioned above, iterative optimization-based methods~\cite{RAFTStereo,CREStereo,igev,crestereo++,dlnr,mochastereo} show impressive results. Following RAFT~\cite{raft}, these methods adopt ConvGRUs~\cite{GRU} to update the disparity map recurrently and bypass \deleted{the} the computational burden of 3D convolutions, making high-resolution stereo matching possible. IGEVStereo~\cite{igev} further provides a better initial disparity map for the updater with a tiny 3D convolution network.  MoCha-Stereo~\cite{mochastereo} achieves accurate detail matching by capturing repeated geometric contours with \added{the} motif~\cite{motif} channel attention mechanism.

\textbf{Cross-Domain Generalization.} \deleted{Existing real-world stereo datasets are too small to adequately train a stereo network that can adapt to diverse scenes.} Stereo networks that can generalize well from synthetic to real domains are promising. \deleted{Cai~\etal~\cite{msstereo} attribute the poor generalization to the learning-based domain-specific features and replace them with traditional feature descriptors.} DSMNet~\cite{DSMNet} designs a domain normalization layer and a non-local graph-based filtering layer to reduce the domain shifts. ITSA~\cite{ITSA} bridges the gap between the extracted features from original and perturbed images to minimize the feature representation sensitivity. GraftNet~\cite{GraftNet} exploits a pre-trained encoder to extract broad-spectrum features and compresses features for better matching. FCNet~\cite{FCNet} introduces a contrastive feature loss and a selective whitening loss to maintain the feature consistency. PCWNet~\cite{PCWNet} fuses the multi-scale volumes to extract domain-invariant features, and narrows the residue searching range for easier matching. Chang~\etal~\cite{hvt} transform the training sample\added{s} into new domains with diverse distributions and minimize the cross-domain feature inconsistency to capture domain-invariant features.

\textbf{\replaced{Ground-truth}{Disparity} Distribution Modeling.} \deleted{Cost volume filtering-based stereo networks~\cite{PSMNet,GwcNet,PCWNet} predict a probability distribution for each pixel, which is naturally suitable for supervision via the cross-entropy loss. However} \added{Unlike classification}, the ground-truth distribution of stereo matching can\deleted{ }not be modeled as \added{a} one-hot \added{vector} due to the conflict between the discrete disparity candidates and the continuous disparity label. PDSNet~\cite{PDSNet} proposes to model the ground-truth distribution\added{s} as discrete Laplacian\added{s with fixed width} and infer the final disparity with the maximum a posterior\added{i} \deleted{(MAP)} estimator. Chen~\etal~\cite{SMNet} model the distribution as the discrete Gaussian and determine the scope of the major mode based on monotonicity during inference. \added{AcfNet~\cite{AcfNet} models uncertainty by computing the variances of the output distributions.} Xu~\etal~\cite{adl} achieve \deleted{the} outstanding \deleted{fine-tuning and} generalization performance by embedding edge information into the ground-truth modeling and optimizing the selection strategy of the major mode. \deleted{In addition, there are some works that model the output distribution as continuous. By predicting \replaced{an}{a real-value} offset for each disparity candidate, CDN~\cite{CDN} turns the output to a mixture of Dirac delta functions. SMDNet~\cite{SMDNet} represents the output distribution as the bi-modal mixture densities. Liu~\etal~\cite{stereorisk} interpolate the discrete distribution \replaced{using}{via} a Laplacian kernel and minimize the risk function to obtain the optimal disparity.} Following these studies, our \replaced{goal}{pursuit} is to \replaced{develop}{model} a better \deleted{multi-modal} ground-truth \added{modeling} for stereo matching.

\textbf{\replaced{Knowledge Distillation}{Dark Knowledge}.} The concept of \replaced{knowledge distillation}{dark knowledge} was first proposed by Hinton~\etal in \cite{hinton2015distilling}. The core idea is to use the \replaced{dark knowledge}{output multi-hot distribution} (\ie, soft labels) \replaced{from}{of} a teacher \replaced{network}{model} to train a student \replaced{network}{model}. By leveraging the soft labels, the student \replaced{network}{model} gains insights into the \replaced{similarity}{relationships} between classes, leading to improved generalization on unseen domain\added{s}. Subsequently, the concept of dark knowledge \replaced{has}{have} been continuously enriched, including attention map\added{s}~\cite{dk_attention}, activation boundary~\cite{dk_activation_boundaries}, relationship graph~\cite{dk_graph},~\etc. \replaced{Additionally}{In addition}, the student can access a wider range of knowledge by aggregating the outputs from \replaced{a network ensemble}{ensemble models} ~\cite{dk_ensemble1,dk_ensemble2,dk_ensemble3}. Despite the widespread use of dark knowledge in classification and segmentation, it \replaced{remains largely unexplored}{has not yet been fully explored} in stereo matching. In this paper, we focus on extracting dark knowledge from \replaced{the pre-trained}{outputs of} ensemble\deleted{ model} to construct multi-modal probability distributions to \replaced{train a domain generalized}{supervise} stereo network\deleted{ training}.


\section{Method}

\begin{figure*}[t]
    \centering
    \includegraphics[width=0.98\linewidth]{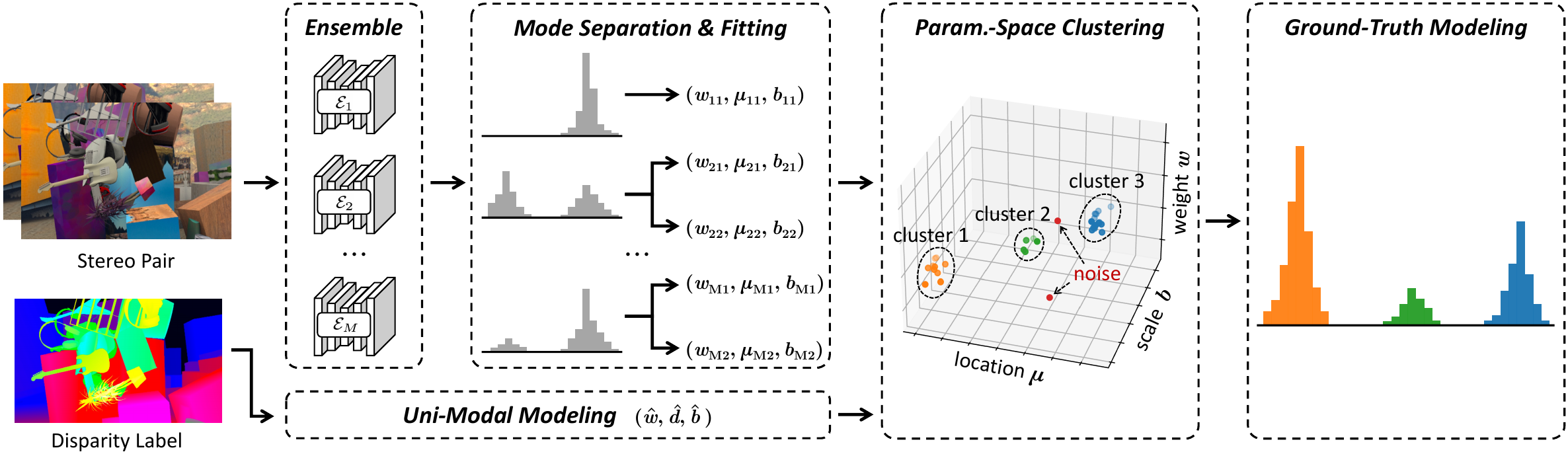}
    \caption{\textbf{Illustration of the ground-truth distribution modeling.} For each pixel in the input image pair, the \replaced{network ensemble predicts}{ensemble models output} $M$ multi-modal probability distributions. Individual modes are separated from these distributions and fitted as parameterized Laplacians $(w,\mu,b)$. The disparity label is also modeled as the uni-modal Laplacian with coordinate $(\hat{w}, \hat{d}, \hat{b})$. Then, we cluster these modes in the parameter space to distinguish \replaced{the objective knowledge (effective clusters) from the biased knowledge (noise)}{the clusters (objective knowledge) from the noise (biased knowledge)}. The elements within each cluster are fused and modeled as a formulated mode in the final ground-truth distribution.}
    \label{GTDM}

    \vspace{-6pt}

\end{figure*}

\subsection{Preliminary}

\replaced{Stereo matching networks based on the cost volume filtering strategy}{Cost volume filtering-based stereo matching networks}~\cite{GCNet,PSMNet,GwcNet,PCWNet} output a discrete probability distribution $\mathbf{p} = [p_0, p_1, \ldots, p_{D-1}] \in \mathbb{R}^{1 \times D}$ for each pixel \replaced{of}{in} the target image, where $D$ is the \added{pre-defined} disparity search range. Treating stereo matching as a classification task~\cite{PDSNet,SMDNet,adl,AcfNet}, the output distribution can naturally be supervised by the cross-entropy loss:

\begin{equation}
\mathcal{L}_{ce}(\mathbf{p},\mathbf{\hat{p}})= -\sum_{d=0}^{D-1} \hat{p}_d \cdot {\rm log}\ p_d \label{cross entropy}
\end{equation}

\noindent where $\mathbf{\hat{p}} = [\hat{p}_0, \hat{p}_1, \ldots,\hat{p}_{D-1}]$ is the ground-truth \deleted{probability} distribution. The core problem \replaced{is}{lies in} that $\mathbf{\hat{p}}$ \replaced{remains}{is} unknown, \replaced{as}{since} the disparity labels $\hat{d}$ \replaced{in}{of} stereo matching are continuous scalars obtained \replaced{from}{by} depth sensors such as LiDAR. \replaced{Several studies}{Some works}~\cite{PDSNet,SMNet} model the ground-truth distribution as a \deleted{formulated} uni-modal distribution, of which the mean value is set to the disparity label. Later, Xu~\etal~\cite{adl} \replaced{demonstrate}{prove} that the multi-modal distribution is more natural than the na\"{i}ve uni-modal distribution at the edge and can provide better supervision signals. \replaced{Considering the lack of exploration of non-edge regions in existing research, we aim to develop a reasonable and effective ground-truth modeling method for all regions in stereo matching.}{This motivates us to explore a more reasonable and effective modeling of the ground-truth distribution across all regions for stereo matching.}

\added{Manually matching similar textures between stereo image pairs to model multi-modal distributions for non-edge regions is exhausting and infeasible.}
We observe that stereo networks typically output \deleted{the} multi-modal distributions \deleted{in} not only \added{in} edge regions but also \added{in} repetitive-textured or transparent regions, and \replaced{produce}{output the} wider \replaced{modes}{distributions} in weak-textured regions, as shown in~\cref{distribution visualization}. \deleted{The} \replaced{M}{m}ultiple modes indicate \deleted{that the pixel in the target image has} multiple similar matches in the reference image, while \deleted{the} wider mode\added{s} \replaced{reflect greater matching uncertainty}{indicates that the match is more uncertain}. \replaced{Therefore, we leverage these two types of dark knowledge to model ground-truth distributions for both edge and non-edge regions.}{Thus, we aim to exploit both the pattern similarity and the matching uncertainty information to model the ground-truth distribution,} \replaced{Our method enriches the information within the ground-truth distribution, enabling}{believing this can help the} stereo network\added{s to} learn more knowledge from the training samples\deleted{ and benefit the cross-domain generalization}.


\subsection{MIDAS}
\label{Ground-Truth Distribution Modeling}

In this section, we introduce how to extract dark knowledge from the pre-trained stereo networks, and leverage it to construct our multi-modal ground-truth distribution.


As shown in~\cref{GTDM}, \added{given a stereo image pair,} the pre-trained \replaced{network ensemble}{ensemble networks} $\{\mathcal{E}_1,\mathcal{E}_2,\ldots,\mathcal{E}_M\}$ \deleted{input a stereo image pair and} outputs a set of probability distributions $\{\mathbf{p}_1,\mathbf{p}_2,\ldots,\mathbf{p}_M\}$ for each pixel in the target image. Then, we separate modes from the output distribution, and fit each mode as a discrete Laplacian:

\vspace{-8pt}
\begin{equation}
    {\rm Laplacian}(\mathbf{d};w,\mu,b) = w \cdot \frac{\mathrm{exp}(-\frac{|\mathbf{d}-\mu|}{b})}{\sum_{d\in\mathbf{d}}\mathrm{exp}(-\frac{|d-\mu|}{b})}
    \label{eq Laplacian}
\end{equation}

\noindent where $\mathbf{d} = [0,1,\ldots,D-1]$ is the disparity candidates. Specifically, the weight parameter $w$ is defined as the sum of the probabilities within the mode. We apply soft-argmin~\cite{GCNet} and mean absolute deviation (MAD)~\cite{MAD} operations upon the normalized mode to compute the location parameter $\mu$ and the scale parameter $ b$ in~\cref{eq Laplacian}, respectively. \replaced{The details are shown in~\cref{alg 1}.}{Please refer to~\cref{alg 1} for more details.} In this way, we transform the predicted distribution to a set of points $\mathcal{T} = \{(w,\mu,b)\}$ in the Laplace parameter space. The number of elements in $\mathcal{T}$ indicates the number of modes in the distribution. $\mu$ denotes the center location of each mode, representing the potential disparity. $w$ denotes the matching probability of each mode, reflecting the degree of similarity between modes. $b$ denotes the width of each mode, reflecting the strength of the texture information in the corresponding region. To mitigate the impact of all pre-trained networks making incorrect predictions simultaneously, we additionally add a uni-modal Laplacian distribution centered around the disparity label $\hat{d}$, \ie, a point with coordinate $(\hat{w}, \hat{d}, \hat{b})$, to the Laplace parameter space.

\begin{algorithm}[t]
    \small
    \caption{Mode Separation and Fitting}
    \label{alg 1}
    \begin{algorithmic}  
        \STATE \textbf{Input:} probability distribution $\mathbf{p}$
        \STATE \textbf{Input:} threshold $\epsilon>0, \sigma>0$
        \STATE \textbf{Output:} a set of triples $\mathcal{T}$

        \STATE $\mathcal{T} \gets \emptyset$

        \WHILE{$\max(\mathbf{p}) > \epsilon$}
            \STATE $l,r \gets \mathrm{argmax}(\mathbf{p})$  
            \WHILE{$\mathbf{p}[l] - \mathbf{p}[l-1] > \sigma$ \AND $l-1 \geq 0$}
                \STATE $l \gets l-1$
            \ENDWHILE
            \WHILE{$\mathbf{p}[r] - \mathbf{p}[r+1] > \sigma$ \AND $r+1 \leq D-1$}
                \STATE $r \gets r+1$
            \ENDWHILE

            \STATE $w \gets \sum_{d=l}^{r} {\mathbf{p}[d]}$ \setstretch{1.2}
            
            \STATE $\mu \gets \sum_{d=l}^{r}{ (\mathbf{p}[d]/w) \cdot d}$ \setstretch{1.2}

            \STATE $b \gets \sum_{d=l}^{r}{ (\mathbf{p}[d]/w) \cdot |d - \mu| }$ \setstretch{1.2}

            \STATE $\mathcal{T} \gets \mathcal{T} \cup \{(w,\mu,b)\}$

            \STATE $\mathbf{p}[l:r] \gets 0$

        \ENDWHILE

        \STATE \textbf{Return} $\mathcal{T}$ 

    \end{algorithmic}
\end{algorithm}

Next, we apply the DBScan algorithm~\cite{dbscan} to perform clustering along the $\mu$-axis, resulting in $K$ clusters $\{C_1,C_2,\ldots,C_K\}$ and a set of noise. Note that even when the label point $(\hat{w}, \hat{d}, \hat{b})$ has no adjacent points, we still consider it as an effective cluster rather than noise. The points within each cluster have similar location parameters, indicating that a significant number of pre-trained networks respond to the similar disparity candidates. We believe that these points are reliable and represent the objective knowledge within the ensemble\deleted{ models}. Conversely, the noisy points correspond to the unreliable modes, which represent the biased knowledge. We remove this biased knowledge during modeling to generate better ground-truth distributions.

Finally, the stereo ground-truth distribution is modeled as a mixture of $K$ Laplacians:


\vspace{-5pt}
\begin{equation}
    \begin{aligned}
    \mathbf{\hat{p}} &= \sum_{k=1}^{K}{\rm Laplacian}(\mathbf{d}; w_k,\mu_k,b_k) \\
     &= \sum_{k=1}^{K}w_k \cdot \frac{\mathrm{exp}(-\frac{|\mathbf{d}-\mu_k|}{b_k})}{\sum_{d\in\mathbf{d}}\mathrm{exp}(-\frac{|d-\mu_k|}{b_k})}
     \end{aligned}
    \label{multi-modal modeling}
\end{equation}

\noindent where $w_k$, $\mu_k$, and $b_k$ are the means of the weight, location, and scale parameters of elements in cluster $C_k$, respectively. Assuming that the label point is located in cluster $C_1$, we set $\mu_1=\hat{d}$. In addition, the modeled distribution is normalized as $\mathbf{\hat{p}} / \|\mathbf{\hat{p}}\|_1$ to ensure that the sum of probabilities is equal to one.


Our method \replaced{offers}{owns} the following advantages: 1) By clustering in the parameter space and filtering out noise, we effectively avoid introducing biased knowledge into the \added{modeled} ground-truth\deleted{ modeling}. 2) Directly superimposing the output distributions of the \added{pre-trained} ensemble\deleted{ model} may result in multiple peaks for \replaced{the fused}{each} mode due to the misalignment of \deleted{the} mode centers. In contrast, fusing modes in the parameter space and remodeling them as the Laplacian can maintain the uni-modal nature\deleted{ of each mode}. 3) Our method can adaptively adjust the mode number in the ground-truth distribution as well as the height and width of each mode not only for edge regions but also for non-edge regions.

\begin{figure}[t]
    \centering
     \includegraphics[width=1.0\linewidth]{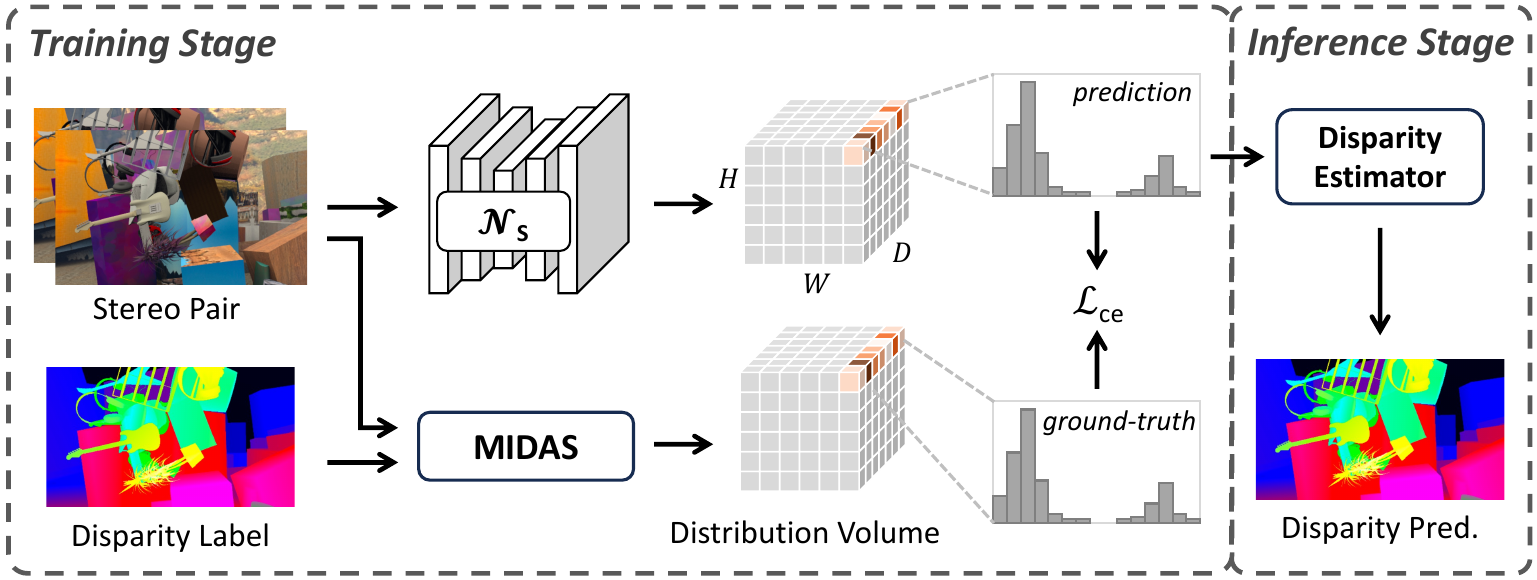}
    \caption{\textbf{Training and inference pipeline\deleted{ of our method}.} $\mathcal{N}_s$: the stereo network to be trained. \replaced{MIDAS}{DKM}: \deleted{dark knowledge based} \added{the} ground-truth distribution modeling \added{method described} in~\cref{Ground-Truth Distribution Modeling}.\deleted{ DME: dominant-modal disparity estimator~\cite{adl}.}}
    \label{pipeline}

\end{figure}

\subsection{Overall Pipeline}

\cref{pipeline} illustrates the overall pipeline of our method.
To train \replaced{a}{the} stereo network $\mathcal{N}_s$, the dark knowledge embedded multi-modal distribution\added{s} described in~\cref{Ground-Truth Distribution Modeling} \replaced{are}{is} applied as the ground-truth \deleted{distributions} to supervise the \replaced{predicted}{output} distributions of $\mathcal{N}_s$.\deleted{Note that the time spent on modeling the ground-truth distribution is negligible, as it is only executed once at the beginning of training and the results are stored for future use.} \added{Following~\cite{depthanything}, we apply color jittering and random occlusion only to the image pairs input to $\mathcal{N}_s$ to encourage the trained network to acquire domain-invariant features.}
Cross-entropy loss encourages the stereo network to output multi-modal distributions~\cite{adl}. Directly applying the weighted average operation (soft-argmin)~\cite{GCNet} to regress disparity on the multi-modal distribution\added{s} \deleted{can} lead\added{s} to over-smoothing artifacts~\cite{SMNet},~\ie the disparity is incorrectly estimated between modes. For this reason, we \replaced{employ}{refer to} DME~\cite{adl} instead of soft-argmin to estimate the final disparit\replaced{ies}{y} from the \replaced{predicted}{disparity} distribution\added{s during inference}.

\begin{table*}[t]
    \small
    \centering
    \setlength{\tabcolsep}{10pt}
    
    \begin{tabular}{lcccccc}
    \toprule
    \multirow{2}{*}{Method}  & \multirow{2}{*}{Publication} & KITTI 2015 & KITTI 2012 & Middlebury & ETH3D & Mean \\
    & & \textgreater 3px & \textgreater 3px & \textgreater 2px & \textgreater 1px   & Rank  \\
    \midrule
    PSMNet~\cite{PSMNet} & CVPR 2018 & 16.30\textsuperscript{\textcolor{red}{18}} & 15.10\textsuperscript{\textcolor{red}{18}} & 25.10\textsuperscript{\textcolor{red}{18}} & 23.80\textsuperscript{\textcolor{red}{18}} &  {18.00} \\
    
    GwcNet~\cite{GwcNet} & CVPR 2018 & 12.80\textsuperscript{\textcolor{red}{17}} & 11.70\textsuperscript{\textcolor{red}{17}} & 18.10\textsuperscript{\textcolor{red}{16}} & 9.00\textsuperscript{\textcolor{red}{16}} &  {16.50} \\

    GANet~\cite{GANet} & CVPR 2019 & 11.70\textsuperscript{\textcolor{red}{16}} & 10.10\textsuperscript{\textcolor{red}{16}} & 20.30\textsuperscript{\textcolor{red}{17}} & 14.10\textsuperscript{\textcolor{red}{17}} & 16.5\\
    
    
    DSMNet~\cite{DSMNet}  & ECCV  2020 & 6.50\textsuperscript{\textcolor{red}{15}} & 6.20\textsuperscript{\textcolor{red}{15}} & 13.80\textsuperscript{\textcolor{red}{13}} & 6.20\textsuperscript{\textcolor{red}{14}} &  {14.25} \\

    CFNet~\cite{CFNet}  & CVPR 2021 & 5.80\textsuperscript{\textcolor{red}{12}} & 4.70\textsuperscript{\textcolor{red}{11}} & 15.30\textsuperscript{\textcolor{red}{14}} & 5.80\textsuperscript{\textcolor{red}{12}} &  {12.25} \\
    
    Mask-CFNet~\cite{maskedstereo} & CVPR 2023 & 5.80\textsuperscript{\textcolor{red}{12}} & 4.80\textsuperscript{\textcolor{red}{12}} & 13.70\textsuperscript{\textcolor{red}{12}} & 5.70\textsuperscript{\textcolor{red}{11}} &  {11.75} \\
    
    Raft-Stereo~\cite{RAFTStereo} & 3DV 2021 & 5.70\textsuperscript{\textcolor{red}{11}} & 5.20\textsuperscript{\textcolor{red}{14}} & 12.60\textsuperscript{\textcolor{red}{11}} & 3.30\textsuperscript{\textcolor{red}{6}} &  {10.50} \\
    
    FC-GANet~\cite{FCNet}  & CVPR 2022 & 5.30\textsuperscript{\textcolor{red}{9}} & 4.60\textsuperscript{\textcolor{red}{10}} & 10.20\textsuperscript{\textcolor{red}{9}} & 5.80\textsuperscript{\textcolor{red}{12}} &  {10.00} \\
    
    PCWNet~\cite{PCWNet} &  ECCV 2022 & 5.60\textsuperscript{\textcolor{red}{10}} & 4.20\textsuperscript{\textcolor{red}{5}} & 15.77\textsuperscript{\textcolor{red}{15}} & 5.20\textsuperscript{\textcolor{red}{10}} &  {10.00} \\
    
    IGEV-Stereo~\cite{igev}  & CVPR 2023 & 6.03\textsuperscript{\textcolor{red}{14}} & 5.18\textsuperscript{\textcolor{red}{13}} & 7.27\textsuperscript{\textcolor{red}{3}} & 3.60\textsuperscript{\textcolor{red}{7}} &  {9.25} \\


    
    Graft-GANet~\cite{GraftNet} & CVPR 2022 &  4.90\textsuperscript{\textcolor{red}{6}} & 4.20\textsuperscript{\textcolor{red}{5}} & 9.80\textsuperscript{\textcolor{red}{8}} & 6.20\textsuperscript{\textcolor{red}{14}} &  {8.25} \\

    ITSA-CFNet~\cite{ITSA}  & CVPR 2022 & 4.70\textsuperscript{\textcolor{red}{4}} & 4.20\textsuperscript{\textcolor{red}{5}} & 10.40\textsuperscript{\textcolor{red}{10}} & 5.10\textsuperscript{\textcolor{red}{9}} &  {7.00} \\
    
    StereoRisk~\cite{stereorisk} & ICML 2024 & 5.19\textsuperscript{\textcolor{red}{8}} & 4.43\textsuperscript{\textcolor{red}{9}} & 9.32\textsuperscript{\textcolor{red}{7}} &\cellcolor{yellow!20} 2.41\textsuperscript{\textcolor{red}{2}} &  {6.50} \\
    
    
    NMRF~\cite{nmrf}  & CVPR 2024 & 5.10\textsuperscript{\textcolor{red}{7}} & 4.20\textsuperscript{\textcolor{red}{5}} & 7.50\textsuperscript{\textcolor{red}{4}} & 3.80\textsuperscript{\textcolor{red}{8}} &  {6.00} \\
    
    GANet + ADL~\cite{adl}  & CVPR 2024 &  4.84\textsuperscript{\textcolor{red}{5}} & 3.93\textsuperscript{\textcolor{red}{4}} & 8.72\textsuperscript{\textcolor{red}{6}} & \cellcolor{red!20} 2.31\textsuperscript{\textcolor{red}{1}} &  {4.00} \\

    

    \midrule
    PSMNet + Ours  & ------ & 4.49\textsuperscript{\textcolor{red}{3}} & \cellcolor{yellow!20} 3.72\textsuperscript{\textcolor{red}{2}} & 7.95\textsuperscript{\textcolor{red}{5}} & 3.17\textsuperscript{\textcolor{red}{5}} &  {3.75} \\
    
    GwcNet + Ours  & ------ & \cellcolor{yellow!20} 4.16\textsuperscript{\textcolor{red}{2}} 
    & 3.74\textsuperscript{\textcolor{red}{3}} 
    & \cellcolor{yellow!20} 7.23\textsuperscript{\textcolor{red}{2}} 
    & 2.91\textsuperscript{\textcolor{red}{4}} 
    &  \cellcolor{yellow!20} {2.75} \\

    PCWNet + Ours & ------ 
    & \cellcolor{red!20} \textbf{3.96}\textsuperscript{\textcolor{red}{1}} 
    & \cellcolor{red!20} \textbf{3.57}\textsuperscript{\textcolor{red}{1}} 
    & \cellcolor{red!20} \textbf{7.20}\textsuperscript{\textcolor{red}{1}} 
    & 2.72\textsuperscript{\textcolor{red}{3}} &  \cellcolor{red!20} {\textbf{1.50}} \\

    \bottomrule
    \end{tabular}
    
    \caption{\textbf{\replaced{Quantitative evaluation of c}{C}ross-domain generalization\deleted{ evaluation}.} All \deleted{the} methods are trained on SceneFlow~\cite{SceneFlow} and evaluated on four popular real-world datasets~\cite{KITTI2012,KITTI2015,Middlebury,ETH3D}. \replaced{Mean rank is computed to compare the comprehensive generalization performance.}{We calculate the average ranking of each method on four (or three) datasets to measure its overall performance.} The \colorbox{red!20}{\textbf{best}} and \colorbox{yellow!20}{second best} are marked with colors.}
    \label{cross-domain generalization}
\end{table*}

\section{Experiments}

\subsection{Datasets and Evaluation Metrics}

Following previous works~\cite{DSMNet, ITSA, FCNet}, we train \deleted{the} stereo networks on \deleted{the} SceneFlow \deleted{synthetic dataset}~\cite{SceneFlow} and evaluate \deleted{the} cross-domain generalization performance on the training sets of four real-world datasets: KITTI 2015~\cite{KITTI2015}, KITTI 2012~\cite{KITTI2012}, Middlebury~\cite{Middlebury}, and ETH3D~\cite{ETH3D}.

SceneFlow is a synthetic dataset that includes 35,454 pairs of stereo images for training and 4,370 pairs for testing. We use its \textit{finalpass} \replaced{training set}{version for training}, as it considers motion blur and depth of field, making it more realistic and \replaced{challenging}{difficult compared to the \textit{cleanpass} version}. KITTI 2012 and 2015 are two \deleted{outdoor} driving scene datasets, including 194 and 200 training pairs respectively. Middlebury and ETH3D are \replaced{small datasets, both containing dozens of real image pairs.}{smaller real-world datasets, which separately contain 15 and 27 training pairs.}

We \replaced{employ}{use} $k$px as the \added{performance} metric, which counts the percentage of \added{valid} pixels with an absolute \added{disparity} error greater than the threshold $k$. For KITTI 2015 and 2012, the threshold is set to 3. For ETH3D, the threshold is set to 1. \replaced{For Middlebury}{In particular}, \deleted{Middlebury provides three different resolutions, and} we evaluate on \replaced{its}{the} half resolution training set \deleted{,} and \added{set} the threshold \deleted{is set} to 2.

\subsection{Implementation Details}
\replaced{We perform experiments on PSMNet~\cite{PSMNet} and GwcNet~\cite{GwcNet}, two classic networks frequently employed as backbones, along with PCWNet~\cite{PCWNet}, which is designed for the cross-domain generalization.}{We conduct experiments on three cost volume filtering-based stereo networks, namely PSMNet~\cite{PSMNet}, GwcNet~\cite{GwcNet}, and PCWNet~\cite{PCWNet}, using NVIDIA RTX 4090 GPUs.} \deleted{We implement all networks in PyTorch and use Adam optimizer with $\beta_1 = 0.9$ and $\beta_2 = 0.999$. When generating the ground-truth distributions,} \replaced{We first}{we} pre-train these networks with \deleted{the} uni-modal Laplacian cross-entropy loss~\cite{PDSNet}\replaced{, and}{. Then, we} select three checkpoints from each network \replaced{to form the ensemble for generating the multi-modal ground-truth distributions.}{, resulting a total of nine output distributions to ensemble.} \deleted{In our experiments,} $\hat{w}$ is set to 1 and $\hat{b}$ is set to 0.8. $\epsilon$ and $\sigma$ in~\cref{alg 1} are all set to $1\times10^{-3}$.  The distance threshold and the density threshold in DBScan algorithm~\cite{dbscan} are set to 3 and 2, respectively. \replaced{Then}{Finally}, we train \replaced{these three}{the stereo} networks with our \replaced{modeled ground-truth}{method} from scratch\replaced{.}{on SceneFlow training set for 80 epochs}
\added{For all experiments, we train the stereo networks on a single NVIDIA 4090 GPU for 80 epochs.} \replaced{We apply Adam ($\beta_1 = 0.9$ and $\beta_2 = 0.999$) to optimize the networks and}{ using} \deleted{the} one-cycle \deleted{learning rate} schedule \replaced{($max\_lr=1\times10^{-3}$) to adjust the learning rate.}{with the max learning rate of $1\times10^{-3}$.}

\subsection{Cross-Domain  Evaluation}

\textbf{\replaced{Comparison}{Quantitative comparison} \added{with state-of-the-art}.} As shown in~\cref{cross-domain generalization}, \added{all three backbones~\cite{PSMNet,GwcNet,PCWNet} trained with our method demonstrate promising generalization performance. In particular,} PCWNet~\cite{PCWNet} with our method achieves \replaced{state-of-the-art}{promising} cross-domain generalization performance on most datasets\deleted{, and ranks in the top three on all four datasets}. Compared \replaced{to}{with} the baseline, it \replaced{yields considerable improvements of}{improves by} 29.29\%, 15.00\%, 54.34\%, and 47.69\% on \deleted{the} KITTI 2015~\cite{KITTI2015}, KITTI 2012~\cite{KITTI2012}, Middlebury~\cite{Middlebury}, and ETH3D~\cite{ETH3D}, respectively. \replaced{These results}{This strongly} prove\deleted{s} that \deleted{the} stereo network\added{s} \replaced{supervised by our multi-modal ground-truth distributions}{using our method} can effectively learn \replaced{generalizable matching principles}{insightful matching knowledge} from synthetic data and \replaced{transfer}{generalize} well to \replaced{real-world scenarios}{the real world}.

We further compute the mean rank to compare the comprehensive generalization performance of the listed methods \replaced{across}{in dealing with} diverse \replaced{scenarios}{scenes}. \added{As shown in~\cref{cross-domain generalization},} PCWNet~\cite{PCWNet} with our method \replaced{rank first}{ deservedly ranks $1^{st}$} among all methods, followed by \deleted{the} GwcNet~\cite{GwcNet} and PSMNet with our method. \replaced{The superior comprehensive performance of our method}{This clearly} \replaced{highlights}{indicates} \replaced{its}{the} potential \deleted{of our method} in \replaced{enhancing the reliability}{ensuring the safety} of downstream tasks~\cite{dsgn,baitingming,zhonghanzhi}.

\deleted{\cref{Comparison with existing methods on different baselines} compares our method with the competitors based on the same backbones. It can be seen that our method outperforms the methods~\cite{FCNet,ITSA,GraftNet} that extract domain-invariant features, as well as those~\cite{adl,PDSNet} that model the ground-truth distribution, by a large margin.}

\begin{figure*}
    \centering
    \includegraphics[width=0.95\linewidth]{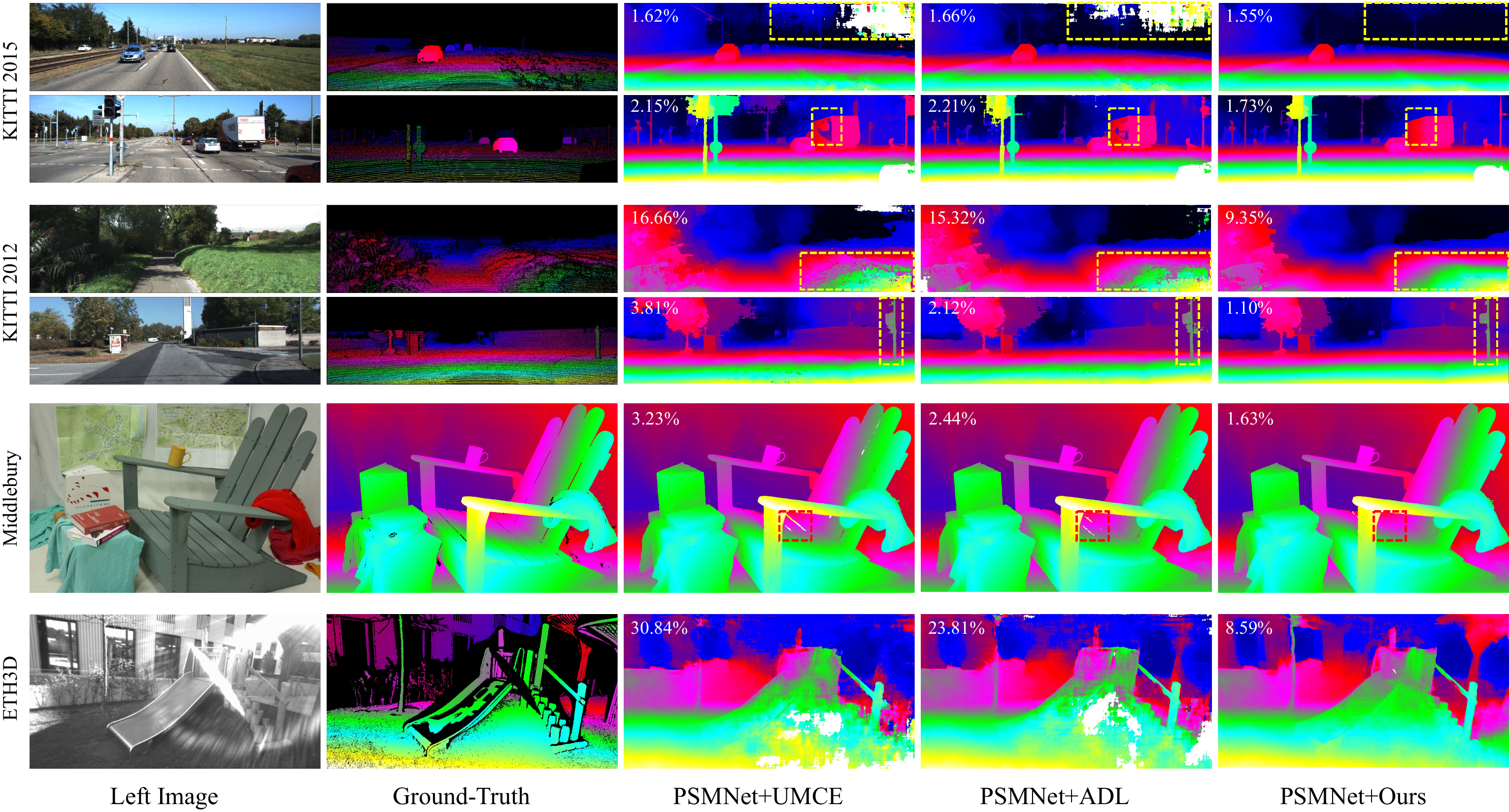}
    \caption{\textbf{\replaced{Qualitative evaluation}{Visualization} of cross-domain generalization.} All \deleted{the} methods are trained on SceneFlow~\cite{SceneFlow} and evaluated on four real-world
datasets~\cite{KITTI2012,KITTI2015,Middlebury,ETH3D}. Compared \replaced{to uni-modal method (UMCE)~\cite{PDSNet} and edge-aware multi-modal method (ADL)~\cite{adl}}{with UMCE~\cite{PDSNet} and ADL~\cite{adl}}, our method \replaced{demonstrates superior performance}{performs better} in weak-textured regions (sky, container), repetitive-textured regions (grass, chair gaps)\added{,} and object edges (sign post). \replaced{Furthermore, o}{O}ur method \deleted{also} exhibits excellent robustness \replaced{when}{in} handling \replaced{strong}{challenging} glare (last row). The outlier metric is displayed in the upper left corner of \replaced{each}{the} disparity map\deleted{s}.}
    \label{Visualization of cross-domain generalization}
\end{figure*}

\textbf{\replaced{Comparison with ground-truth modeling.}{Single checkpoint model comparison.}} \deleted{Most of the methods listed in~\cref{cross-domain generalization} report the best results for each dataset, which usually come from different checkpoints.} In practical applications, \deleted{the} stereo network\added{s} can only load a fixed checkpoint. Therefore, 
\deleted{we are \replaced{particularly}{more} interested in }the generalization performance of \added{a} single checkpoint \added{deserves special attention}. We compare the single checkpoint performance of our method with \replaced{both}{the} uni-modal~\cite{PDSNet} and  multi-modal~\cite{adl} \added{ground-truth modeling} methods. For each method, we \replaced{select the best checkpoint among}{compare the comprehensive performance of} all its checkpoints, following the ranking strategy \replaced{described}{used} in~\cref{cross-domain generalization}. \deleted{The best single checkpoint performance is shown in~\cref{single checkpoint performance}.} \replaced{As shown in~\cref{single checkpoint performance},}{It is encouraging that} our single checkpoint  \replaced{performance (last row)}{model} even surpasses the best performance of \replaced{ADL (third row)~\cite{adl}}{the similar methods}\added{, which models the multi-modal distributions for edge regions only, on all datasets.} We also report the degradation of \added{the} single checkpoint performance compared to the best performance on each dataset. \replaced{The uni-modal method~\cite{PDSNet} and the edge-aware multi-modal method~\cite{adl} degrade by 12.06\% and 5.30\% respectively}{The single checkpoint performance of  uni-modal method degrades severely (12.06\%)}, while our method only degrades by 2.81\%. \replaced{This further demonstrates the advantage of our method that models multi-modal distributions for both edge and non-edge regions.}{This demonstrates that our superior generalization performance is not due to the randomness in the training process, but rather that the stereo network truly learns universal matching capabilities.}

\begin{table}
    \footnotesize
    \centering
    \setlength{\tabcolsep}{1.2pt}

    \begin{tabular}{lccccc}
    \toprule
    \multirow{2}{*}{Method}   & KT15 & KT12 & MB & ETH3D & Average \\
    & \textgreater 3px & \textgreater 3px & \textgreater 2px & \textgreater 1px  & Degradation  \\
    
    \midrule
    UMCE~\cite{PDSNet} & 4.73  & 4.64  & 9.76  & 4.18  &  \\
    UMCE* & 5.62\textsubscript{\textcolor{red}{-18.82\%}}  & 5.55\textsubscript{\textcolor{red}{-19.61\%}}  & 9.76\textsubscript{\textcolor{red}{-0.00\%}}  & 4.59\textsubscript{\textcolor{red}{-9.81\%}}  & -12.06\% \\
    \midrule
    ADL~\cite{adl} & 4.78 & 4.23 & 8.85 & 3.44 &  \\
    ADL* & 4.78\textsubscript{\textcolor{red}{-0.00\%}} & 4.23\textsubscript{\textcolor{red}{-0.00\%}} & 8.95\textsubscript{\textcolor{red}{-1.13\%}} & 4.13\textsubscript{\textcolor{red}{-20.06\%}} & \cellcolor{yellow!20} -5.30\% \\
    \midrule
    Ours & \cellcolor{red!20} \textbf{4.49}  & \cellcolor{red!20} \textbf{3.72}  & \cellcolor{red!20} \textbf{7.95}  & \cellcolor{red!20} \textbf{3.17} &   \\
    Ours* & \cellcolor{red!20} \textbf{4.49}\textsubscript{\textcolor{red}{-0.00\%}}  & \cellcolor{red!20} \textbf{3.72}\textsubscript{\textcolor{red}{-0.00\%}}  & \cellcolor{yellow!20} {8.29}\textsubscript{\textcolor{red}{-4.28\%}}  & \cellcolor{yellow!20} {3.39}\textsubscript{\textcolor{red}{-6.94\%}}  & \cellcolor{red!20} \textbf{-2.81\%}\\
    \bottomrule

    \end{tabular}
    \caption{\textbf{\replaced{Quantitative}{Generalization} comparison  \replaced{with ground-truth modeling methods}{of he PSMNet variants}.} \added{All methods are trained on PSMNet~\cite{PSMNet}.} *represents \deleted{the best single checkpoint} \replaced{generalization performance from the optimal single checkpoint}{model}. We also report the \added{performance} degradation of \added{the} single checkpoint \deleted{performance} \replaced{relative}{compared} to the best \replaced{result}{performance} on each dataset, indicated by red subscripts.}

    \label{single checkpoint performance}

\end{table}

\begin{figure}[t]
    \centering
    \includegraphics[width=0.92\linewidth]{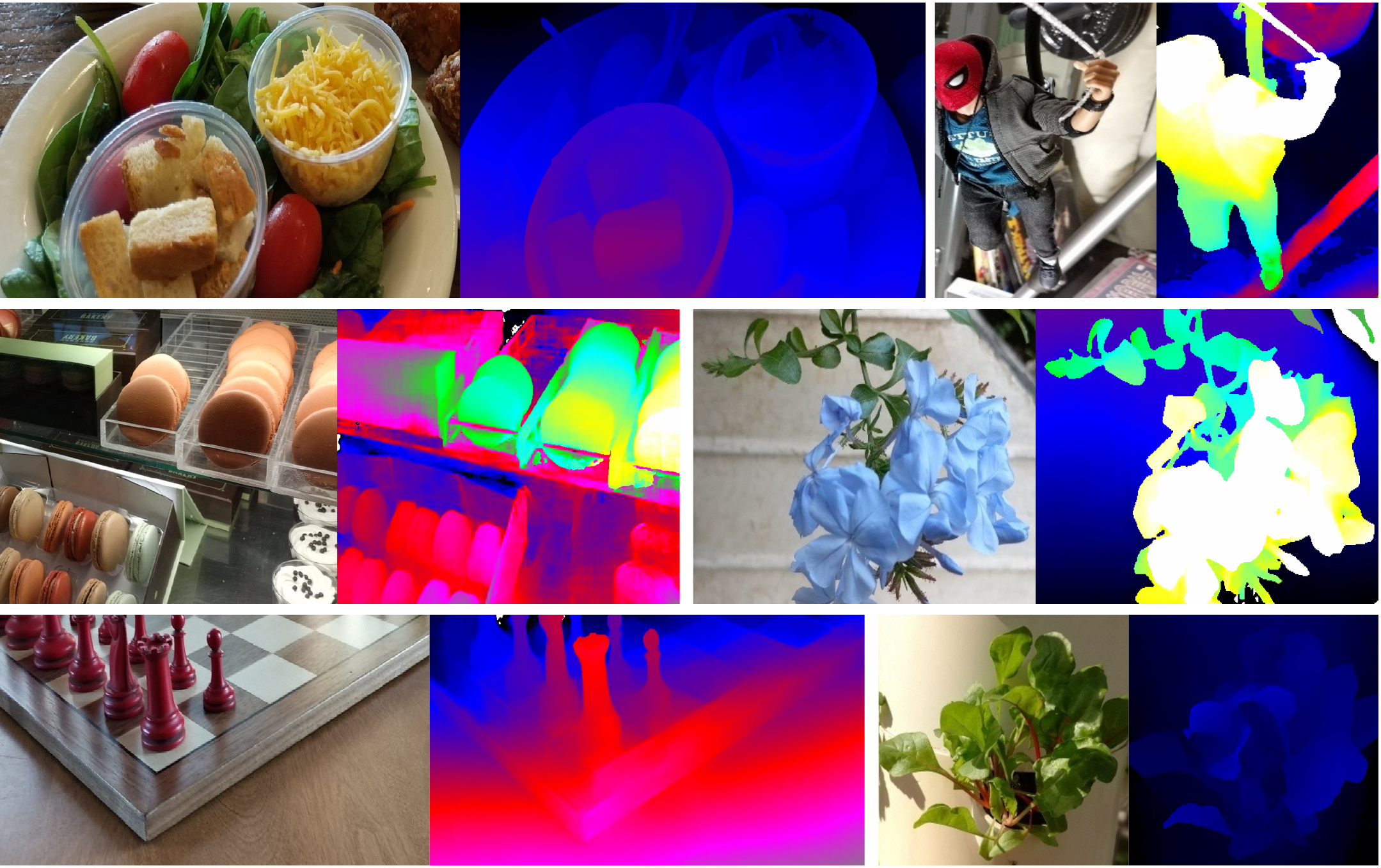}
    \caption{\textbf{Qualitative evaluation of cross-domain generalization} on daily-life images from Holopix50k~\cite{hua2020holopix50k}.}
    \label{daily images}
\end{figure}

\begin{table}[t]
    \small
    \setlength{\tabcolsep}{7pt}
    \centering
    \begin{tabular}{l c c c c }
        \toprule
        Method & KT15 & KT12 & MB & ETH3D \\
        \midrule
        KD~\cite{hinton2015distilling} & 4.67 & 4.41 & 9.00 & 4.02 \\
        \midrule
        Average~\cite{dk_ensemble3} & 4.58 & 3.96 & 8.61 & \cellcolor{yellow!20} {3.53} \\
        Entropy~\cite{dk_ensemble2} & 4.61 & 4.10 & 8.29 & 3.71 \\
        Confidence~\cite{dk_ensemble4} & \cellcolor{yellow!20} {4.54} & \cellcolor{yellow!20} {3.78} & \cellcolor{red!20} \textbf{7.58} & 3.75 \\
        \midrule
        Ours & \cellcolor{red!20} \textbf{4.49} & \cellcolor{red!20} \textbf{3.72} & \cellcolor{yellow!20} {7.95} & \cellcolor{red!20} \textbf{3.17}  \\
        \bottomrule  
    \end{tabular}
    \caption{\textbf{Quantitative comparison with knowledge distillation methods.}  \added{Beside the vanilla distillation~\cite{hinton2015distilling}, we evaluate three ensemble methods that employ average~\cite{dk_ensemble3}, entropy-based~\cite{dk_ensemble2}, and confidence-aware~\cite{dk_ensemble4} fusion strategies, respectively. All methods are trained on PSMNet~\cite{PSMNet}.}}
    \label{ablation: distillation}
\end{table}

\begin{table*}[t]
    \centering
    \small
    
    \setlength{\tabcolsep}{5pt}
    \begin{tabular}{c | c c c c c c | c}
         \toprule
         Method & PSMNet~\cite{PSMNet} & PCWNet~\cite{PCWNet} & GwcNet~\cite{GwcNet} & IGEVStereo~\cite{igev} & ICGNet~\cite{icgnet} & GANet+ADL~\cite{adl} & PCWNet+Ours\\
         \midrule
         \textgreater 3px & 4.56 & 3.68 & 3.30 & 2.47 & 2.34 & \cellcolor{red!20}\textbf{1.81} & \cellcolor{yellow!20} 2.03\\
         \bottomrule
    \end{tabular}
    \caption{\textbf{Quantitative \replaced{evaluation}{results} on SceneFlow test set~\cite{SceneFlow}}. We report the percentage of outliers with the absolute error larger than 3px.}
    \label{sceneflow results}
\end{table*}

We visualize the disparity maps to provide more intuitive comparisons with \replaced{the modeling}{similar} methods. As shown in~\cref{Visualization of cross-domain generalization}, in weak-textured regions such as the sky (top row), both uni-modal~\cite{PDSNet} and multi-modal~\cite{adl} methods result in large areas of artifacts, which are not accurately reflected in the metrics because the sky lacks disparity labels. The same phenomenon can be observed in the white areas of containers (second row). Our method also demonstrates superior performance in handling repetitive-textured regions, achieving smoother disparity for grass (third row) and avoiding mismatching the chair \added{gaps} (fifth row). Furthermore,our method \deleted{can} obtain\added{s} sharp and accurate object edges (fourth row), which is crucial for downstream tasks such as 3D object detection~\cite{dsgn}\replaced{, and}{Finally, it is remarkable that our method} exhibits considerable robustness \replaced{when addressing the challenging case of strong glare}{in handling challenging glare} (last row). These examples demonstrate that by modeling the ground-truth distribution\added{s} as \deleted{the ensembled} multi-modal distribution\added{s for both edge and non-edge regions} and \added{adaptively} adjusting the width of \replaced{each mode}{the modes}, \replaced{our method}{we} successfully guide\added{s} \replaced{stereo}{the} network\added{s} to learn matching similarity and uncertainty to cope with \replaced{real-world}{the} complexities.~\cref{daily images} shows more qualitative results of our method on daily-life images of Holopix50k~\cite{hua2020holopix50k}.

\begin{figure}[t]
    \centering
    \small
    \includegraphics[width=0.88\linewidth]{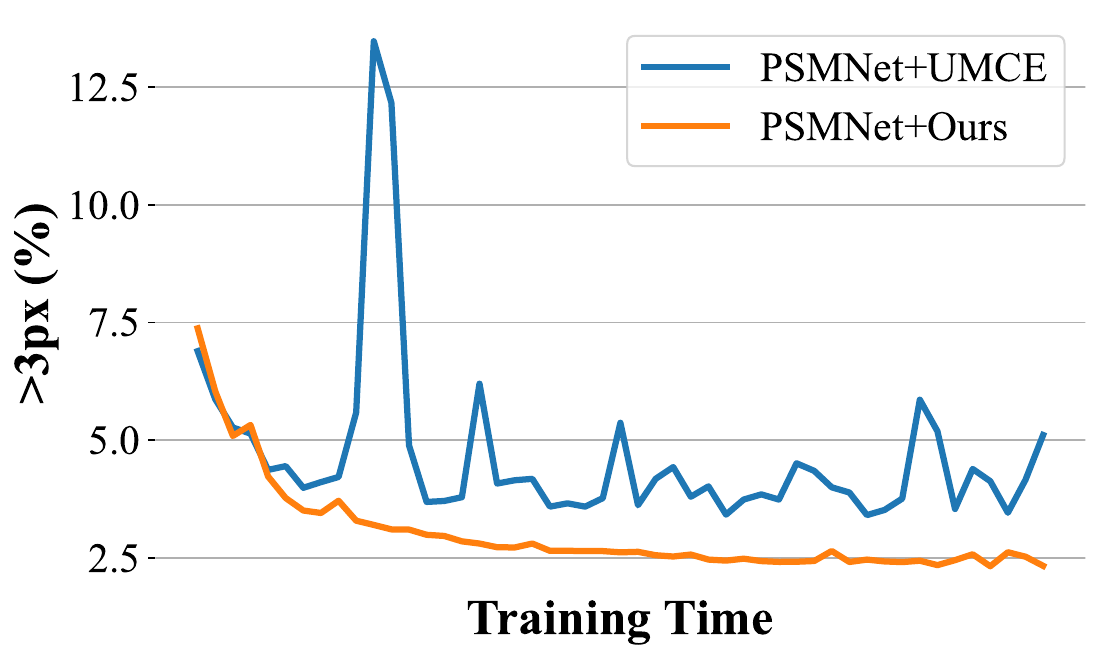}
    \caption{\textbf{Visualization of {outlier rate curves}} \added{during training} on SceneFlow test set~\cite{SceneFlow}.  We report the percentage of outliers with the absolute error larger than 3px.}
    \label{convergence}
\end{figure}

\textbf{\added{Comparison with knowledge distillation.}} \deleted{In this section, we compare our method with {the} knowledge distillation methods.} As shown in~\cref{ablation: distillation}, compared with \replaced{the vanilla knowledge distillation~\cite{hinton2015distilling,xuruoyu}}{KD}, \replaced{ensemble methods~\cite{dk_ensemble3, dk_ensemble4,dk_ensemble2}}{EKD} aggregate\deleted{s} the output distributions of multiple pre-trained networks and effectively improve\deleted{s} the generalization of \replaced{stereo}{the} network\added{s}. Our method achieves further generalization improvement \replaced{on most datasets}{over EKD}, which we attribute to our fusion strategy for dark knowledge. Simply superimposing \replaced{the predicted}{multiple probability} distributions \replaced{ introduces the}{may lead to} noisy \replaced{modes}{peaks} and \replaced{compromises}{destroy} the uni-modal nature of each mode due to the misalignment of the mode centers. We propose to fuse modes in the parameter space and remodel\deleted{ing} \replaced{the aggregated clusters}{them} as Laplacians, \replaced{making}{which makes} the ground-truth distribution \replaced{better}{more} formulated and easier \deleted{for the network} to learn.

\subsection{In-Domain Evaluation}

\textbf{\replaced{Comparison with state-of-the-art.}{Quantitative comparison.}} Although our method \replaced{primarily}{mainly} focuses on the cross-domain generalization, \replaced{it}{we} still achieve\added{s} competitive performance in the \replaced{source}{original} domain~\cite{SceneFlow}, as shown in~\cref{sceneflow results}. PCWNet~\cite{PCWNet} with our method \deleted{only} generates \added{only} 2.03\% outliers, \replaced{representing a 44.84\% improvement over the baseline.}{ranking $2^{nd}$ among all methods.} \replaced{Both ADL~\cite{adl} and our method outperform competitors trained with L1 loss. This is attributed to the finer-grained supervision signals provided by the modeled distributions compared to the original disparity labels, enabling the stereo networks to learn more efficiently.}{We attribute this to the correction of the true mode center using ground-truth label, which can ensure the accuracy of the supervision signal.}

\textbf{Convergence stability.} To further \replaced{investigate}{explore} the impact of different ground-truth distributions on \deleted{the learning of the} stereo network \added{learning}, we plot the outlier rate curves during training on \deleted{the} SceneFlow test set~\cite{SceneFlow}. As illustrated in~\cref{convergence}, our method achieves more stable and better convergence compared to the uni-modal method~\cite{PDSNet}. The stereo network effectively learns valuable dark knowledge from our \replaced{proposed}{modeled} multi-modal ground-truth distributions, thereby \replaced{reducing}{avoiding} overfitting to the training set.


    
    



\begin{table}[t]
    \setlength{\tabcolsep}{6.5pt}
    \small
    \centering
    \begin{tabular}{c c c c c c}
         \toprule
         \#Arch. & \#CKPT & KT15 & KT12 & MB & ETH3D \\
         \hline
         0 &	0 &	4.73 & 4.64 & 9.76 & 4.18 \\
         1 &	1 &	4.67 &	4.41 &	9.00 &	4.02 \\
         \hline
         1 &	2 &	\cellcolor{yellow!20} {4.57} &	3.87 &	8.47 & \cellcolor{yellow!20} {3.34} \\
         2 &	1 & 4.64 & 3.89 & 8.27 & 3.64 \\
         2 & 2 & 4.59 & \cellcolor{yellow!20} {3.82} & \cellcolor{yellow!20} {8.01} & 3.40 \\
         3 & 3 & \cellcolor{red!20} \textbf{4.49} & \cellcolor{red!20} \textbf{3.72} & \cellcolor{red!20} \textbf{7.95} & \cellcolor{red!20} \textbf{3.17} \\
         \bottomrule
    \end{tabular}
    \caption{\textbf{Ablation study} of the number of network architectures and checkpoints in the pre-trained ensemble. All methods are trained on PSMNet~\cite{PSMNet}.}
    \label{ablation}
\end{table}

\begin{table}[t]
    \setlength{\tabcolsep}{6pt}
    \centering
    \small
    \begin{tabular}{lcccc}
    \toprule
    Method & KT15 & KT12 & MB & ETH3D \\
    \midrule
     PSMNet~\cite{PSMNet} + Ours & \cellcolor{red!20} \textbf{4.49} & \cellcolor{red!20} \textbf{3.72} & \cellcolor{red!20} \textbf{7.95} & \cellcolor{red!20} \textbf{3.17}  \\
     w/o BKF & {4.57} & {3.81} & {8.47} & {3.40} \\
    \bottomrule
    \end{tabular}
    \caption{\textbf{Ablation study} of biased knowledge filtering (BKF).}
    \label{ablation: biased knowledge}
\end{table}

\subsection{Ablation Study}

\cref{ablation} presents the ablation study on the number of network architectures and checkpoints in the pre-trained ensemble. Our methods (second block) consistently outperform the modeling method (first row)~\cite{PSMNet} and the distillation method (second row)~\cite{hinton2015distilling}. Compared to ensembling different architectures, ensembling different checkpoints helps mitigate training randomness and achieves better generalization performance. Increasing the number of architectures and checkpoints yields further improvements in generalization. Due to computational resource constraints, we do not ablate larger-scale ensembles, which we believe could still achieve some level of performance improvement.

We also perform an ablation study to verify the effectiveness of filtering biased knowledge. As shown in~\cref{ablation: biased knowledge}, the preservation of noisy points clustered in the Laplace parameter space leads to poorer generalization performance across all datasets, demonstrating that these points are indeed detrimental to network learning.

\section{Conclusion}
\replaced{
In this paper, we present MIDAS, a novel method that extracts similarity and uncertainty information from the pre-trained network to model ground-truth distributions for stereo matching. To mitigate the impact of domain preferences in a single network, we employ the network ensemble. Specifically, we propose to aggregate objective knowledge in the Laplace parameter space and ultimately model the ground-truth distribution as a mixture of Laplacians.}{In this paper, we introduce DKM, a novel multi-modal ground-truth distribution modeling method for stereo matching. Different from the prior works, we highlight and alleviate the cross-domain preferences within the pre-trained stereo networks. To achieve superior cross-domain generalization, we propose to extract and filter the dark knowledge within the ensemble models in Laplace parameter space, and model the ground-truth distribution as a mixture of Laplacians.} Our method  \replaced{effectively improves the generalization performance of}{is generic and effective for} cost volume filtering-based stereo networks. \deleted{PCWNet with our method achieves the state-of-the-art generalization on KITTI 2015 and 2012 , and leads the comprehensive ranking on four popular real-world datasets.} By \deleted{guiding the network to} \replaced{modeling}{learn} \replaced{intuitive patterns for both edge and non-edge regions}{pattern similarity and matching uncertainty}, our method exhibits considerable robustness in handling \added{real-world} challeng\replaced{es}{ing regions}. \replaced{Last but not least, our single checkpoint network achieves stable generalization performance across diverse scenarios, satisfying the practical demands of real-world applications.}{Furthermore, we achieve excellent generalization with single checkpoint model, demonstrating that the trained stereo network indeed learns insightful matching knowledge.}

\noindent \textbf{Acknowledgments.} This work was supported by The Key Research \& Development Plan of Zhejiang Province under Grant No.2024C01017, 2024C01010.

{
    \small
    \bibliographystyle{ieeenat_fullname}
    \bibliography{main}
}


\end{document}